\documentclass[10pt,twocolumn,letterpaper]{article}

\usepackage{xcolor}
\usepackage{graphicx}

\usepackage[font=scriptsize]{caption}

\usepackage[export]{adjustbox}
\usepackage{tabularx}
\usepackage{ulem}
\graphicspath{{./images/}}

\usepackage[pagebackref,breaklinks,colorlinks]{hyperref}

\usepackage[capitalize]{cleveref}
\crefname{section}{Sec.}{Secs.}
\Crefname{section}{Section}{Sections}
\Crefname{table}{Table}{Tables}
\crefname{table}{Tab.}{Tabs.}

\begin{document}

\title{Navigating an Ocean of Video Data: Deep Learning for Humpback Whale Classification in YouTube Videos}


\author{Michelle Ramirez\\
{\tt\small ramirez.528@osu.edu}}
\date{}
\maketitle

\newcommand\mynotes[1]{\textcolor{red}{#1}}

\begin{abstract}

Image analysis technologies empowered by artificial intelligence (AI) have proved images and videos to be an opportune source of data to learn about humpback whale (Megaptera novaeangliae) population sizes and dynamics. With the advent of social media, platforms such as YouTube present an abundance of video data across spatiotemporal contexts documenting humpback whale encounters from users worldwide. In our work, we focus on automating the classification of YouTube videos as relevant or irrelevant based on whether they document a true humpback whale encounter or not via deep learning. We use a CNN-RNN architecture pretrained on the ImageNet dataset for classification of YouTube videos as relevant or irrelevant. We achieve an average  85.7\% accuracy, and 84.7\% (irrelevant)/ 86.6\% (relevant) F1 scores using five-fold cross validation for evaluation on the dataset. We show that deep learning can be used as a time-efficient step to make social media a viable source of image and video data for biodiversity assessments.
\end{abstract}

\section{Introduction}

Humpback whales (\textit{Megaptera novaeangliae}) are faced with the challenge of responding to marine ecosystem changes onset by human-driven climate change and encroachment on marine habitats. With the global ban on commercial whaling, humpback whales are increasing in abundance but continue to face threats from large-scale human activity such as energy production and mining, transportation, biological resource use, and pollution \cite{IUCN}. The unprecedented rates of human-driven large-scale threats places increased importance on data-driven species monitoring efforts to assess population sizes and dynamics of key species such as humpback whales \cite{ScalingBiodiversityMonitoring, DecliningCalvingRates}. 
However, the data gathered by research teams to inform data-driven population assessments typically reports insights on a much smaller geographical scale than the wide-ranging regions humpback whales traverse and provides limited coverage over biologically significant areas \cite{Flukebook}. 

Recent shifts to use image analysis technologies coupled with artificial intelligence (AI) for species population assessments has allowed conservation teams to overcome the challenges attributed to the geographical scope of data collection methods as well as the resources and time dedicated to traditional monitoring methods \cite{ Wildbook, MLConservationPerspectives, AINaturalists}. Flukebook (https://www.flukebook.org/)  addresses limitations in data-collection methods for cetacean populations, including humpback whales, by using species-specific machine learning algorithms to identify and catalogue individuals via images \cite{Flukebook}.  With powerful computational tools available for scalable species monitoring such as Flukebook, the challenge becomes one of data availability \cite{ScalingBiodiversityMonitoring}. 

Social media is an abundant source of image and video data, and has the potential to provide a scope of biodiversity monitoring data otherwise unfeasible to conservation teams \cite{Flukebook, MLConservationPerspectives, SocialMediaDolphins, SocialMediaConservationMethods}. With an expansive global distribution of users contributing data worldwide, images and videos extracted from social media platforms make for an opportune source of documented humpback whale sightings that would otherwise go unnoticed by conservation teams. 
One of the caveats of images and videos retrieved via social media, however, is the variance in relevancy of images and videos contributed under a common search term. Text queries made to social media application programming interfaces (APIs) tend to retrieve massive amounts of relevant results, but also a substantial amount of irrelevant results depending on the user's use case. Filtering through the massive amounts of data retrieved from social media to ensure that the image or video is relevant to species monitoring contexts is a tedious task for human annotators and impedes further work in assessing species populations.

As a first step towards converting social media platforms into a more efficient source of data that documents humpback whale encounters, we present a deep learning approach to classify publicly-accessible YouTube videos retrieved via the YouTube Data API v3 as relevant or irrelevant.  In our context, we define relevant videos as those containing true humpback whale encounters and irrelevant videos otherwise. We use a CNN-RNN deep learning model to perform classification and automate the separation of relevant from irrelevant videos among the total results retrieved from a species-specific text query made to the YouTube Data v3 API.  In the following sections, we outline previous work on using video classification in species monitoring contexts, our data collection and pre-processing methods, the CNN-RNN deep learning architecture used for classification, and results from our study. We conclude by presenting limitations and future work to motivate the use of social media as a source of wildlife monitoring data. 

\section{Related work}

With computational tools available for large-scale data processing, several domains have benefited from the application of tools such as deep learning to solve complex domain-specific challenges. In this section, we cover related work in the application of deep learning for video classification, and outline existing work that has applied video classification in biodiversity monitoring contexts. 

\subsection{Video classification}

Common supervised video classification methods include two-stream convolutional neural networks (CNNs), 3D CNNs, recurrent neural networks (RNNs), and I3D models \cite{ren2019survey}. Since videos are sequences of images, several video classification techniques use a CNN-RNN architecture to capture the temporal order of video frames. In video feature fusion approaches, CNNs are applied as a first step to extract features from video frames. The CNN output is then fed into an RNN often consisting of Long Short-Term Memory (LSTM) cells for time series modeling \cite{ren2019survey, yuedeep}. Using a CNN-RNN approach enables learning a global description of the video's temporal evolution for accurate video classification. 

In the context of biodiversity monitoring, video classification is a powerful approach to automate the processing of camera trap content. Animal detection and action recognition has been studied using deep learning image and video classification techniques \cite{SchindlerCameraTrapVideoClassification, BeeryCameraTrapVideoClassification}. Further work has been done to improve the performance of deep learning when applied to camera traps, and increase the performance granularity to being able to detect, count, and identify animals at the species-level \cite{BeeryCameraTrapVideoClassification, FishSpeciesVideoClassification}.  Video classification techniques for species-level classification are commonly bound to camera trap and satellite imagery datasets \cite{BeeryCameraTrapVideoClassification, MLConservationPerspectives, ScalingBiodiversityMonitoring}, but limited work has been done to apply similar techniques on social media image and video datasets \cite{MLConservationPerspectives}. 

\section{Methods}

We propose a deep learning pipeline to classify publicly-accessible YouTube (www.youtube.com) videos as relevant or irrelevant as a first step towards automating and making social media a more time-efficient and spatially-expansive source of species monitoring data on humpback whales. In this study, we focus on automating classification of YouTube videos for humpback whales via a CNN-RNN deep learning model. In our context, we define relevant videos as those that contain a true humpback whale encounter and irrelevant videos as those otherwise. In this section, we cover the data collection via the YouTube Data application programming interface (API) v3 (developers.google.com/youtube/v3), data pre-processing, and the CNN-RNN deep learning model used for classification in further detail. 

\subsection{Data collection}

We collected a total of 407 publicly-accessible YouTube videos with the YouTube Data API v3 using a general text query of ``humpback whale” to retrieve both relevant and irrelevant results. To maintain balanced relevant and irrelevant classes, we collect 203 relevant videos and 204 irrelevant videos. The YouTube videos were manually labeled by searching through each video for evidence of a humpback whale encounter (relevant) or lack thereof (irrelevant).We follow the ethical and responsible guidelines for social media data collection as outlined in \cite{SocialMediaDataGovernanceGuidelines}. No individually identifying user information was retained and all video titles were renamed to a standardized format. 

\subsection{Data preparation}

For each relevant video, we manually identify a 10-20 second interval where a humpback whale is evidently present. The length of occurrence intervals collected per relevant video varies but is maintained in the 10-20 second range to fully capture the scope of a humpback whale encounter. For irrelevant videos, we randomly select a 15 second interval to maintain consistency in the amount of content extracted from each video. 

\begin{figure}
\centering
\includegraphics[width=0.99\linewidth]{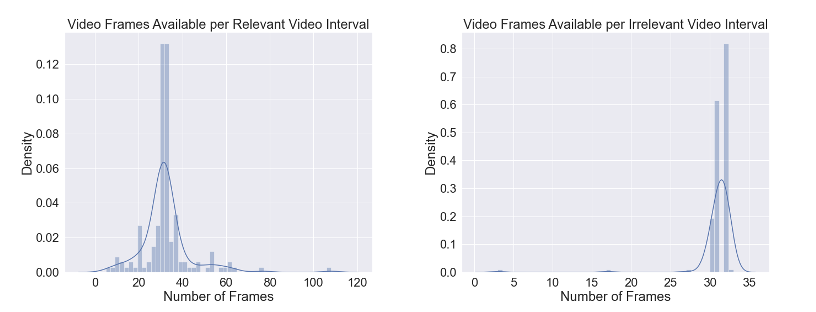}
\captionof{figure}{Distribution plots of frames available across relevant and irrelevant video intervals. The average number of frames available is approximately 31 for both relevant and irrelevant video intervals.}
\label{fig:frame_hist}
\end{figure}

Across all relevant and irrelevant video intervals, we extract 31 video frames per video to standardize the amount of content gathered across the entire video dataset. The number of video frames extracted was determined by averaging the total number of frames available across relevant and irrelevant video intervals, which averaged to 31 (\cref{fig:frame_hist}). Videos with intervals that yielded less than 31 video frames had the middle frame replicated in-place to meet the 31 video-frame count while keeping the original ordering of frames.

\subsection{CNN-RNN model}

The first component of our deep learning model is using a convolutional neural network (CNN) architecture to extract features from the video frames. We resize all of the video frames to the required input size of 224x224 pixels as required by the Inception V3 deep learning model (keras.io/api/applications/inceptionv3/). We take advantage of transfer learning in this first step of our deep learning model by using the Inception V3 image recognition model base pretrained with ImageNet weights and average pooling to extract features from the video frames. 

\begin{figure}
\centering
\includegraphics[width=\linewidth]{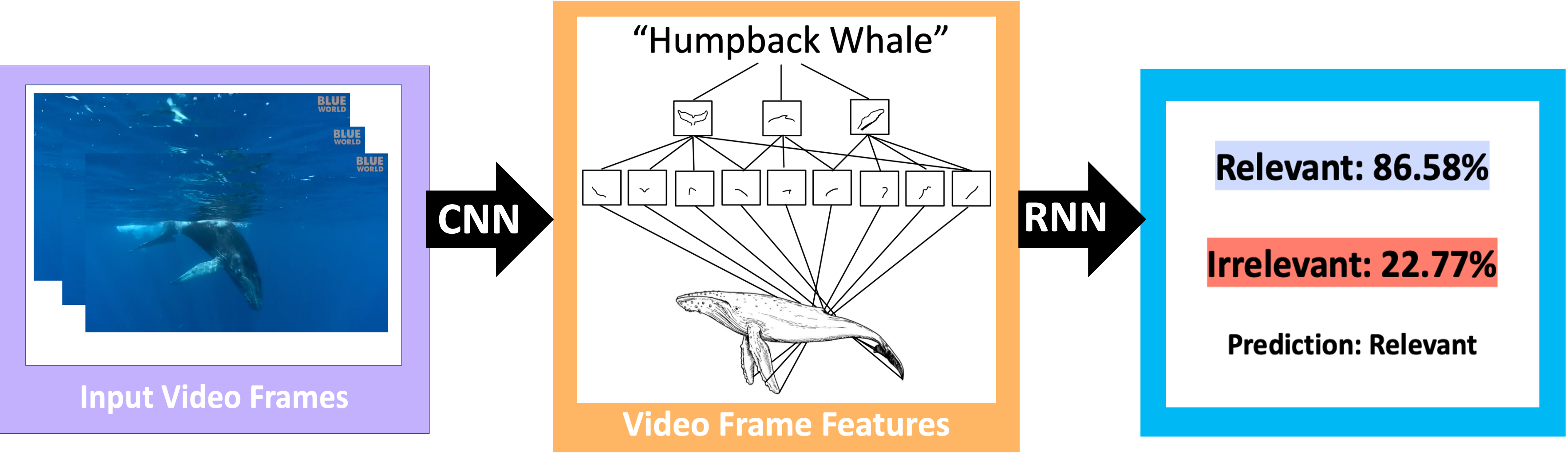}
\captionof{figure}{CNN-RNN model pipeline for humpback whale classification in YouTube videos.}
\label{fig:pipeline_diagram}
\end{figure}

\quad In the second step of our deep learning model, we feed the extracted features obtained via Inception V3 through a subsequent recurrent neural network (RNN) to obtain a final relevance classification on the YouTube video. The RNN used for classification consists of two gated recurrent unit (GRU) layers, a dropout layer with dropout rate of 0.4, a dense layer with rectified linear unit (ReLU) activation, and a final dense layer with SoftMax activation. The RNN model is compiled and trained over the dataset using sparse categorical crossentropy loss. The final classification of a YouTube video's relevance status is made by the RNN, assigning the class (relevant or irrelevant) predicted with greater confidence to the video (\cref{fig:pipeline_diagram}).

\section{Results}

To assess the performance of our CNN-RNN deep learning model, we used five-fold cross validation to partition our dataset into train and test videos. Accuracy, precision, recall, and F1 scores were collected at each fold and averaged to obtain a comprehensive evaluation of our deep learning model's performance at classifying YouTube videos in the defined context of relevancy status. Our CNN-RNN model achieved an average 85.7\% accuracy and in order of irrelevant/relevant classes, a 91.3/82.0 average precision, 79.5/92.1 average recall, and 84.7/86.6 average F1 scores (\cref{tab:model_metrics}). 

\begin{table} 
\centering
\resizebox{\linewidth}{!}{%
\begin{tabular}{|c|c|c|c|c|}
\hline 
Fold & Accuracy & Precision & Recall & F1 Score \\
\hline
1 & 89.0 & 92.1/86.4 & 85.4/92.7 & 88.7/89.4 \\
\hline
2 & 87.8 & 97.0/81.6 & 78.0/97.6 & 86.5/88.9\\
\hline
3 & 86.4 & 85.4/87.5 & 87.5/85.4 & 86.4/86.4 \\
\hline
4 & 84.0 & 88.9/80.0 & 78.0/90.0 & 83.1/84.7 \\
\hline
5 & 81.5 & 93.3/74.5 & 68.3/95.0 & 78.9/83.5 \\
\hline
\bf{Average} & \bf{85.7} & \bf{91.3/82.0} & \bf{79.5/92.1} & \bf{84.7/86.6} \\
\hline
\end{tabular}}
\caption{CNN-RNN model performance metrics for classification task of labeling YouTube videos as relevant or irrelevant. Precision, Recall, and F1 Scores are given in irrelevant/relevant order.}
\label{tab:model_metrics}
\end{table}

\subsection{Correctly classified YouTube videos}

\begin{figure*}
\centering
\includegraphics[width=0.84\linewidth]{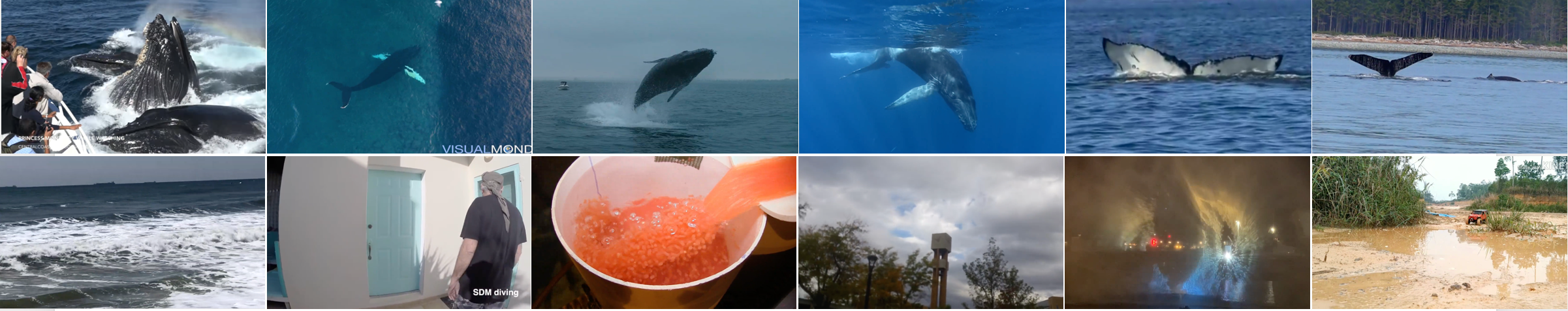}
\caption{Top row: sample frames of videos correctly predicted as relevant. From left to right, top row: humpback whales lunge feeding, drone footage of a humpback whale, humpback whale breaching, underwater footage of a humpback whale, humpback whale fluke, footage of two humpback whales (dorsal fin and fluke). Bottom row: sample frames of videos correctly predicted as irrelevant. From left to right, bottom row: ocean footage (no humpback whales), beach trip vlog, salmon hatchery, personal vlog, evening light show, mud driving. All relevant and irrelevant videos were retrieved by the same query ("humpback whale") made to the YouTube Data v3 API.}
\label{fig:correct_predictions}
\end{figure*}

Our model performed well at predicting both relevant and irrelevant classes. The CNN-RNN model was able to correctly classify relevant videos documenting humpback whale encounters captured from various angles and behaviors. The correctly classified relevant videos included footage of visible whale flukes, breaching, lunging, underwater encounters, and aerial views captured via drone footage (\cref{fig:correct_predictions}). Correctly classified irrelevant videos consisted of varied video content lacking humpback whale encounters. Some irrelevant videos contained content overlapping with oceanic backgrounds and habitats, such as scuba diving videos (\cref{fig:correct_predictions}). Furthermore, we note that YouTube attracts an audience of varying photographer expertise (amateur to expert), therefore videos are subject to vary in quality (duration, resolution, features of humpback whale captured). Nonetheless, the deep learning model was able to perform well on classifying the relevance status of the YouTube videos.

\subsection{Misclassified YouTube videos}

\begin{figure*}
\centering
\includegraphics[width=0.84\linewidth]{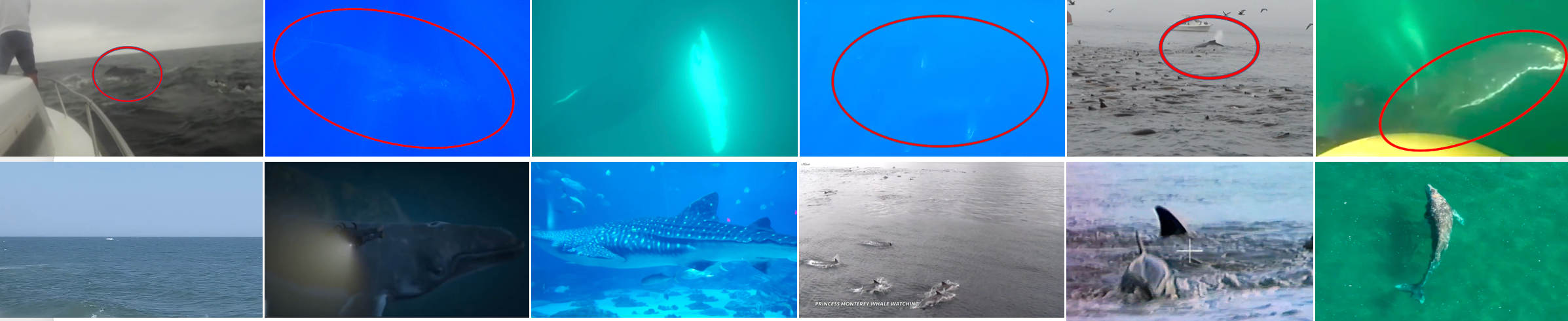}
\caption{Top row: relevant videos that were falsely predicted irrelevant. Humpback whales are hard to identify due to blending in with background, cloudy water conditions, or video resolution. Humpback whales are circled in red to assist the reader in identifying species occurrence. Bottom row: irrelevant videos that were falsely predicted relevant by the CNN-RNN model. Video contents from left to right, bottom row: video of an animal surfacing far off in the distance (cannot identify species), a humpback whale video game, whale shark at an aquarium, a pod of dolphins, footage of killer whales, drone footage of a gray whale.}
\label{fig:incorrect_predictions}
\end{figure*}

Relevant videos that were misclassified as irrelevant mainly arose due to issues of poor visibility (cloudy water, or humpback whale blending in with background/environment) impeding the identification of a humpback whale in the video (Figure 4). Irrelevant videos that were misclassified as relevant consisted mainly of videos that included other marine species but not humpback whales (ex. gray whales, whale sharks, killer whales) and humpback whale video games (\cref{fig:incorrect_predictions}).\newline

\subsection{Further improvement}

Our work demonstrates the potential to significantly reduce the time required of a human annotator to isolate biologically relevant videos from the massive amounts of data retrieved via YouTube for species monitoring assessments.

As this was the first architecture proposed for this case study, we have not extensively
explored the space of parameters, data augmentation techniques, and deep learning architectures possible for our classification task. We outline key points to consider for future improvement in the model's performance based on our findings:
\begin{enumerate}
    \item Including more irrelevant videos of other marine species that closely resemble humpback whales for improved species-level classification.
    \item Cropping video frames to avoid training on the background of videos and place a stronger emphasis on classification by species \cite{BeeryCameraTrapVideoClassification}.
    \item Training the deep learning model with full-length YouTube videos instead of isolating each video to 10-20 second intervals.
    \item Training different deep learning model architectures and comparing the performance(s) against the performance of the CNN-RNN model applied in this work. 
\end{enumerate}

\section{Conclusion}

The unprecedented rates of biodiversity loss present an urgency to scale computational tools for species monitoring and assessments. Our work motivates the consideration of social media as an additional source of data for biodiversity monitoring by using deep learning to address the time-consuming annotation process of massive amounts of social media video data. We present the results of using a CNN-RNN deep learning model to automate annotation of relevant videos for humpback whale encounters. In our work, we show that deep learning performs with an average 85.7\% accuracy in automating the distinction between relevant and irrelevant videos retrieved under a shared text search query from social media platforms. 

We motivate the consideration and investigation of underlying biases such as encounter coverage influenced by geographic hotspots of human presence, areas of high bandwidth, species novelty, and charismatic species biases when sourcing image and video data from social media for species assessments. Social media should be considered in conjunction to other data sources such as fieldwork, camera traps, and satellite imagery to gain a comprehensive understanding of the current state of biodiversity and mitigate the biases in data obtained from social media platforms \cite{ScalingBiodiversityMonitoring}. 

In the digital age, image and video data is one of the most accessible and scalable sources of biodiversity data to inform conservation policy with due diligence. In this study, we focused on using deep learning to find relevant YouTube videos documenting humpback whale encounters but we would like to emphasize that this approach could be adapted and extended to capture insights on additional vulnerable or data-deficient species. We hope our work illuminates the potential of using social media as a source of data for both individualistic and large-scale species monitoring assessments. 

\bibliographystyle{plain}
\bibliography{references}

\end{document}